\documentclass[conference]{IEEEtran}
\IEEEoverridecommandlockouts

\usepackage{bibnames}
\usepackage{amsmath,amssymb,amsfonts}
\usepackage{algorithmic}
\usepackage{graphicx}
\usepackage{textcomp}
\usepackage{makecell}

\usepackage{bm}
\usepackage[table,xcdraw]{xcolor}
\usepackage{array}
\usepackage{booktabs}
\usepackage{tabularx}
\usepackage{siunitx}
\usepackage{caption}
\usepackage{float}
\usepackage{scalerel}
\usepackage{tikz}
\usepackage{graphicx}
\usepackage{subfig}
\usepackage{tikz}
\usepackage{nicematrix}
\usepackage{balance} 
\usepackage{multirow}
\usepackage{mathtools}
\usetikzlibrary{svg.path}

\def\BibTeX{{\rm B\kern-.05em{\sc i\kern-.025em b}\kern-.08em
    T\kern-.1667em\lower.7ex\hbox{E}\kern-.125emX}}

\makeatletter

\newcommand{\linebreakand}{%
  \end{@IEEEauthorhalign}
  \hfill\mbox{}\par
  \mbox{}\hfill\begin{@IEEEauthorhalign}
}
\makeatother

\definecolor{orcidlogocol}{HTML}{A6CE39}
\tikzset{
  orcidlogo/.pic={
    \fill[orcidlogocol] svg{M256,128c0,70.7-57.3,128-128,128C57.3,256,0,198.7,0,128C0,57.3,57.3,0,128,0C198.7,0,256,57.3,256,128z};
    \fill[white] svg{M86.3,186.2H70.9V79.1h15.4v48.4V186.2z}
                 svg{M108.9,79.1h41.6c39.6,0,57,28.3,57,53.6c0,27.5-21.5,53.6-56.8,53.6h-41.8V79.1z M124.3,172.4h24.5c34.9,0,42.9-26.5,42.9-39.7c0-21.5-13.7-39.7-43.7-39.7h-23.7V172.4z}
                 svg{M88.7,56.8c0,5.5-4.5,10.1-10.1,10.1c-5.6,0-10.1-4.6-10.1-10.1c0-5.6,4.5-10.1,10.1-10.1C84.2,46.7,88.7,51.3,88.7,56.8z};
  }
}

\newcommand\orcidicon[1]{\href{https://orcid.org/#1}{\mbox{\scalerel*{
\begin{tikzpicture}[yscale=-1,transform shape]
\pic{orcidlogo};
\end{tikzpicture}
}{|}}}}

\usepackage{hyperref} 

\begin{document}

\title{Autonomous Robotic Radio Source Localization via a Novel Gaussian Mixture Filtering Approach
\thanks{The research was carried out at the Jet Propulsion Laboratory, California Institute of Technology, under a contract with the National Aeronautics and Space Administration (80NM0018D0004).}
\thanks{\copyright 2025. California Institute of Technology. All rights reserved.}
}

\author{\IEEEauthorblockN{Sukkeun Kim \orcidicon{0000-0001-6903-5437}}
\IEEEauthorblockA{\textit{Faculty of Engineering and Applied Sciences} \\
\textit{Cranfield University}\\
Cranfield, UK}
\IEEEauthorblockA{\textit{Jet Propulsion Laboratory} \\
\textit{California Institute of Technology}\\
Pasadena, USA}

\and

\IEEEauthorblockN{Sangwoo Moon \orcidicon{0000-0002-4467-4096}}
\IEEEauthorblockA{\textit{Jet Propulsion Laboratory} \\
\textit{California Institute of Technology}\\
Pasadena, USA}

\and

\IEEEauthorblockN{Ivan Petrunin \orcidicon{0000-0002-8705-5308}}
\IEEEauthorblockA{\textit{Faculty of Engineering and Applied Sciences} \\
\textit{Cranfield University}\\
Cranfield, UK}

\linebreakand

\IEEEauthorblockN{Hyo-Sang Shin \orcidicon{0000-0001-9938-0370}}
\IEEEauthorblockA{\textit{Faculty of Engineering and Applied Sciences} \\
\textit{Cranfield University}\\
Cranfield, UK}
\IEEEauthorblockA{\textit{Cho Chun Shik Graduate School of Mobility} \\
\textit{Korea Advanced Institute of Science and Technology}\\
Daejeon, Republic of Korea}

\and

\IEEEauthorblockN{Shehryar Khattak \orcidicon{0000-0002-9304-1455}}
\IEEEauthorblockA{\textit{Jet Propulsion Laboratory} \\
\textit{California Institute of Technology}\\
Pasadena, USA}
}

\maketitle

\begin{abstract}
This study proposes a new Gaussian Mixture Filter (GMF) to improve the estimation performance for the autonomous robotic radio signal source search and localization problem in unknown environments. The proposed filter is first tested with a benchmark numerical problem to validate the performance with other state-of-the-practice approaches such as Particle Filter (PF) and Particle Gaussian Mixture (PGM) filters. Then the proposed approach is tested and compared against PF and PGM filters in real-world robotic field experiments to validate its impact for real-world applications. The considered real-world scenarios have partial observability with the range-only measurement and uncertainty with the measurement model. The results show that the proposed filter can handle this partial observability effectively whilst showing improved performance compared to PF, reducing the computation requirements while demonstrating improved robustness over compared techniques.
\end{abstract}


\section{Introduction}
Autonomous robotic radio signal source search and localization is a problem in which the robot actively searches for the radio signal source in an unknown environment. Denniston \textit{et al.}~\cite{Denniston_Signal} proposed a graph model for robotic radio signal source search which can be scaled to a large number of measurements. Azuma \textit{et al.}~\cite{Azuma_Stocahstic} proposed a controller for the signal source search in the completely unknown space based on the simultaneous-perturbation stochastic approximation algorithm. Li \textit{et al.}~\cite{Li_Multi} proposed and tested cooperative multi-robot source search problem with gradient descent and Turgemen and Werner~\cite{Turgeman_Multi} proposed Multiple Extrema Search Algorithm (MESA) for source search with multiple agents.

\begin{figure}[t]
  \centering
  \includegraphics[scale=0.27]{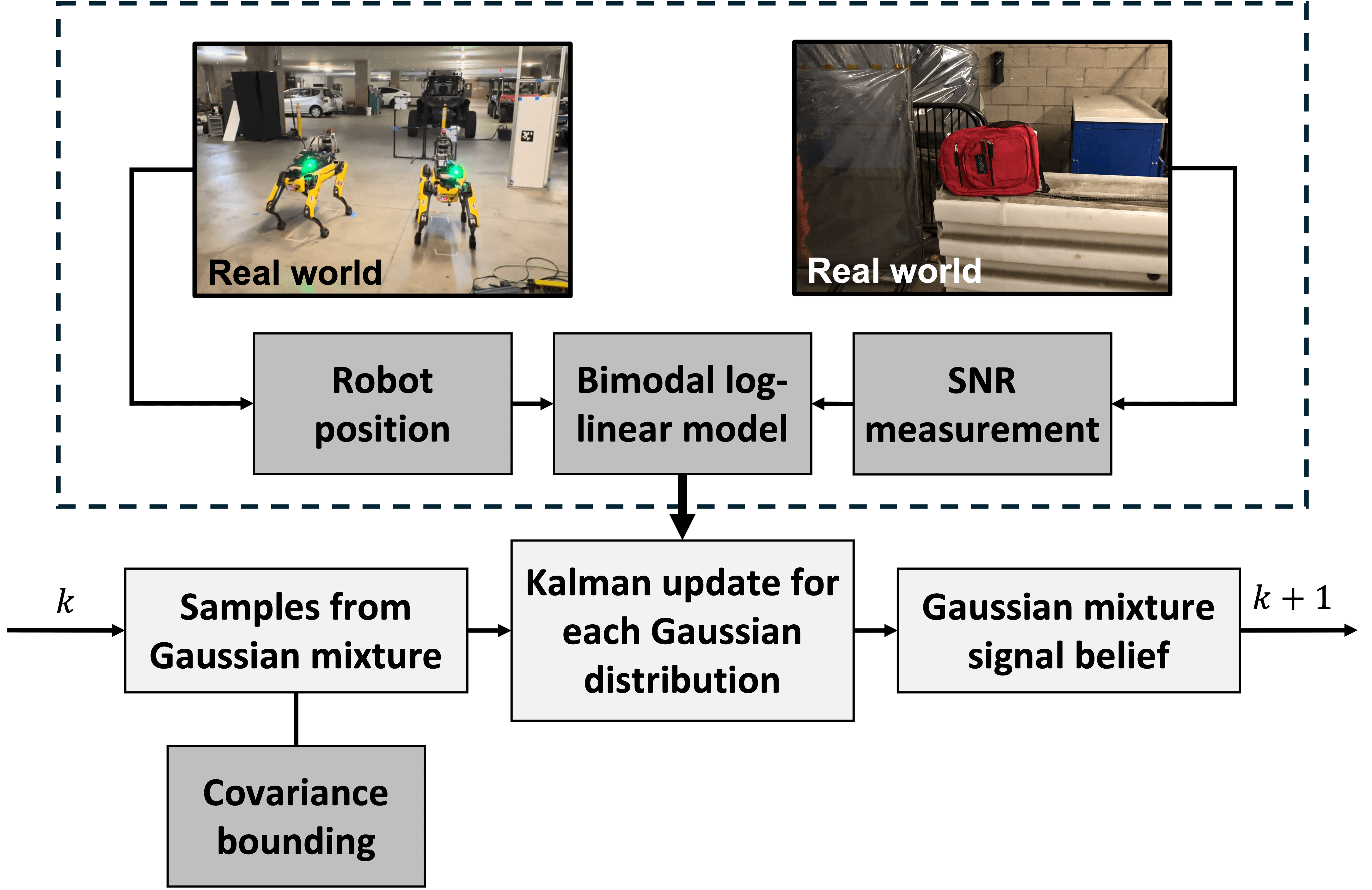}
  \caption{\textbf{Overview of approach.} At the given time instance, the filter takes the real-world robot position and SNR measurement as input to update the filter by calculating the predicted signal strength with bimodal log-linear model. The filter then outputs Gaussian mixture signal belief.}
  \label{fig:overview}
\end{figure}

However, partial observability is one of the biggest issues in the robotic radio signal source search problem due to either no prior knowledge regarding the search environment or due to the limitation of the sensor \cite{Agha_Nebula, Park_source}. This issue can be compensated by the incremental exploration of the unknown environment by the robot; however, the radio signal search problem can become intractable if the robot conducts a naive exploration of the unknown space without leveraging the radio signal source information to guide its search even if the environment size is within the robots operational capabilities. To alleviate this issue, multiple estimation algorithms were considered for the autonomous robotic signal source search. Clark \textit{et al.}~\cite{Clark_Propem} proposed Propagation Environment Modelling and Learning (PropEm-L) to enhance the radio signal by extracting Line-of-Sight (LOS) information. However, this requires an understanding of maps and geometry. To remove these constraints and apply it to real-time applications, estimation filters such as Extended Kalman Filter (EKF) or Particle Filter (PF) ~\cite{Gordon_PF, Djuric_PF, Candy_BPF} are considered. However, PFs suffer from the curse of dimensionality and the performance of the PF can be significantly degraded if the diversity of the particles is lost, typically referred to as the particle degeneracy and impoverishment problem~\cite{Li_PFReivew}. To alleviate these issues, Bayes filters with Gaussian Mixture Model (GMM) have been proposed. However, Gaussian Sum Filter (GSF) \cite{Alspach_GSF}, which is the original idea of Gaussian Mixture Filter (GMF), and its variants have an issue of covariance collapse in complex scenarios. Another approach, the Particle Gaussian Mixture (PGM) filter and its variant \cite{Raihan_PGM1,Raihan_PGM2, Kim_PGMDS} resolved this issue but required high computation. Furthermore, the PGM filters are not able to approximate the ring-shaped signal belief in the range-only measurement scenario which is the case for the considered radio source localization problem. In this study, we propose a novel GMF for the autonomous robotic radio signal source search and localization problem. The proposed filter considers each particle as a Gaussian component to approximate the posterior distribution and is equipped with Matrix Inequality (MI) bounding to set the upper limit of the covariance of each component and uses a scaled sample covariance for resampling. These points distinguish the proposed filter from previous ones. Then it was compared with the state-of-practice PF and PGM filters. The overview of the proposed approach is shown in Fig. \ref{fig:overview}. The main contributions of this study are listed as follows:
\begin{enumerate}
\item A new Gaussian Mixture Filter (GMF) with MI bounding and scaled sample covariance is proposed for autonomous robotic radio signal source search and localization in unknown environments.
\item The proposed filter is validated using a numerical simulation and tested in real-world robotic scenarios using a team of legged NeBula-Spot robots.
\item Comparisons with multiple state-of-the-art nonlinear filters are conducted to provide a comprehensive analysis.
\end{enumerate}

The rest of this paper is composed as follows: Section \ref{pre} briefly introduces the PF and PGM filters. Section \ref{GMF} introduces the proposed GMF approach and Section \ref{Numerical} presents the simulation result of the proposed method on a benchmark numerical example. Section \ref{Scenario} demonstrates the real-world application of the proposed approach for conducting autonomous robotic radio signal source search experiments in complex unknown environments, followed by the discussion. Finally, we conclude this study with future work in Section \ref{conclusion}.~\looseness=-1

\section{Preliminaries}\label{pre}
\subsection{Particle Filter}\label{PF}
PF is a sample-based recursive Bayesian estimation in which the samples approximate the posterior distribution of the state. The general form of the Probability Density Function (PDF) of the recursive Bayesian estimation can be calculated as follows:

\begin{equation}
  p(\bm{x}_k|\bm{z}_{1:k}) = \frac{p(\bm{z}_k|\bm{x}_k)p(\bm{x}_k|\bm{z}_{1:k-1})}{\int p(\bm{z}_k|\bm{x}_{k})p(\bm{x}_{k}|\bm{z}_{1:k-1})d\bm{x}_{k}}, \label{eq1}
\end{equation}

\noindent where the subscript $k = 1,2,3 \dots$ implies the time instance, $\bm{x}_k$ is the state vector and $\bm{z}_{k}$ is the measurement vector. However, an analytical solution of Eq. (\ref{eq1}) is not obtainable for most cases. PF approximates Eq. (\ref{eq1}) with samples and Dirac delta function as follows: 

\begin{equation}
  p(\bm{x}_k|\bm{z}_{1:k}) \simeq \sum^{N_p}_{i=1} w_i \delta(\bm{x}_k-\bm{x}^i_k), \label{eq2}
\end{equation}

\noindent where $N_p$ is the total number of particles, $w_i$ is the weight of $i^{th}$ particle which sums up to 1, $\delta(\cdot)$ is the Dirac delta function and $\bm{x}^i$ is the belief state of the $i^{th}$ particle. A simple Sequential Importance Resampling (SIR) is utilized for comparison in this study.

\subsection{Particle Gaussian Mixture Filter}\label{PGM}
PGM filters are the extension of PF bringing the GMM into the PF to improve the performance of PF. PGM filters were originally proposed to overcome the particle degeneracy issue by approximating the PDF of the state as a GMM. In order to approximate the PDF with GMM, the particles are first clustered with modified K-means clustering~\cite{Raihan_PGM1, Raihan_PGM2} or Density-based Spatial Clustering of Applications with Noise (DBSCAN)~\cite{Kim_PGMDS}, and each cluster is considered as a Gaussian component. A total number of mixture in the GMM is $N_m$ and the GMM can be calculated as follows:

\begin{equation}
  GMM(\bm{x}) = \sum_{n=1}^{N_m} w_n \mathcal{N}(\bm{\mu}_n,\,\bm{\sigma}_n^{2}), \label{eq3}
\end{equation}

\noindent where $w_n$ is the weight of the $n^{th}$ Gaussian component which sums up to 1 and $\bm{\mu}_n$ and $\bm{\sigma}_n^{2}$ are mean and variance of $n^{th}$ Gaussian component, respectively. The new particles for the next time step are sampled from the GMM in Eq.~\eqref{eq3} during the resampling step. In this study, we compare the PGM filter with \textbf{\underline{D}}B\textbf{\underline{S}}CAN (PGM-DS) and the PGM filter with \textbf{\underline{D}}BSCAN and \textbf{\underline{U}}nscented Transform (UT) (PGM-DU). Further details about these filters can be found in Table~\ref{tab:comparison} and in \cite{Kim_PGMDS}.

\begin{table*}[h]
\caption{Comparison of PF, PGM filters and proposed filter.}
\centering
\setlength{\tabcolsep}{5pt}
\begin{tabularx}{\textwidth}{@{}p{3.2cm}X X X X@{}}
\toprule
\textbf{Step} & \textbf{PF} & \textbf{PGM filters} & \textbf{Proposed} \\
\midrule
\textbf{Initialization} & 
Random samples from Gaussian distribution. \newline No assumed covariance. & 
Random samples from Gaussian distribution. \newline No assumed covariance. & Random samples from Gaussian distribution. \newline Assumed covariance with bandwidth parameter. \\

\textbf{Time propagation} & 
Using dynamics directly. \newline No propagation of covariance. & 
Using dynamics directly. \newline No propagation of covariance. & Using dynamics directly. \newline Covariance propagation with Jacobian. \\

\textbf{Clustering} & 
- & 
Particle clustering for update. \newline Sample covariance of each cluster. & - \\

\textbf{MI bounding} & 
- & 
- & Covariance bounding with scaled sample covariance. \\

\textbf{Measurement update} & 
Weight update with measurement likelihood. & Update like EKF with sample covariance (but no covariance update). \newline Weight update with measurement likelihood. & Update like EKF with covariance from the previous step (with covariance update). \newline Weight update with measurement likelihood. \\

\textbf{Resampling} & 
Resampling with weight and number of effective particles. & Resample from GMM. & Resample from GMM and reset the covariance with scaled sample covariance. \\
\bottomrule
\end{tabularx}
\label{tab:comparison}
\end{table*}

\section{Gaussian Mixture Filter}\label{GMF}
The proposed GMF and PGM filters both assume the GMM for their PDF of the state but differ in the approach for the GMM construction. Instead of clustering the particles to form the Gaussian distribution and reconstructing the GMM with these Gaussian components, the proposed GMF considers each particle as a Gaussian distribution. Next, each Gaussian component is propagated directly with nonlinear dynamics and updated in the same manner as it is done in an EKF. However, the uncertainty can be increased after the time propagation and this causes the increase of the covariance of each particle. The increment of individual covariance can lead to greater overlap between the covariances of adjacent Gaussian components, which degrades the ability to represent the unknown PDF. This is a common issue and Psiaki \cite{Psiaki_GaussianSum, Psiaki_Blob} proposed the MI bounding to set the upper limit of the covariance of each Gaussian component. The proposed filter is equipped with the MI bounding and this limits the expansion of covariance of the Gaussian component over the limit and ensures the GMM is closer to the unknown distribution we are trying to approximate using GMM. Furthermore, the proposed filter utilizes the concept of scaled sample covariance and this characteristic differentiates the proposed filter from other GMFs such as \cite{Psiaki_Blob}. \looseness=-1

\begin{figure*}[t!]
\centering
    \subfloat[Approximation of PF.]{\includegraphics[clip,width=0.33\linewidth]{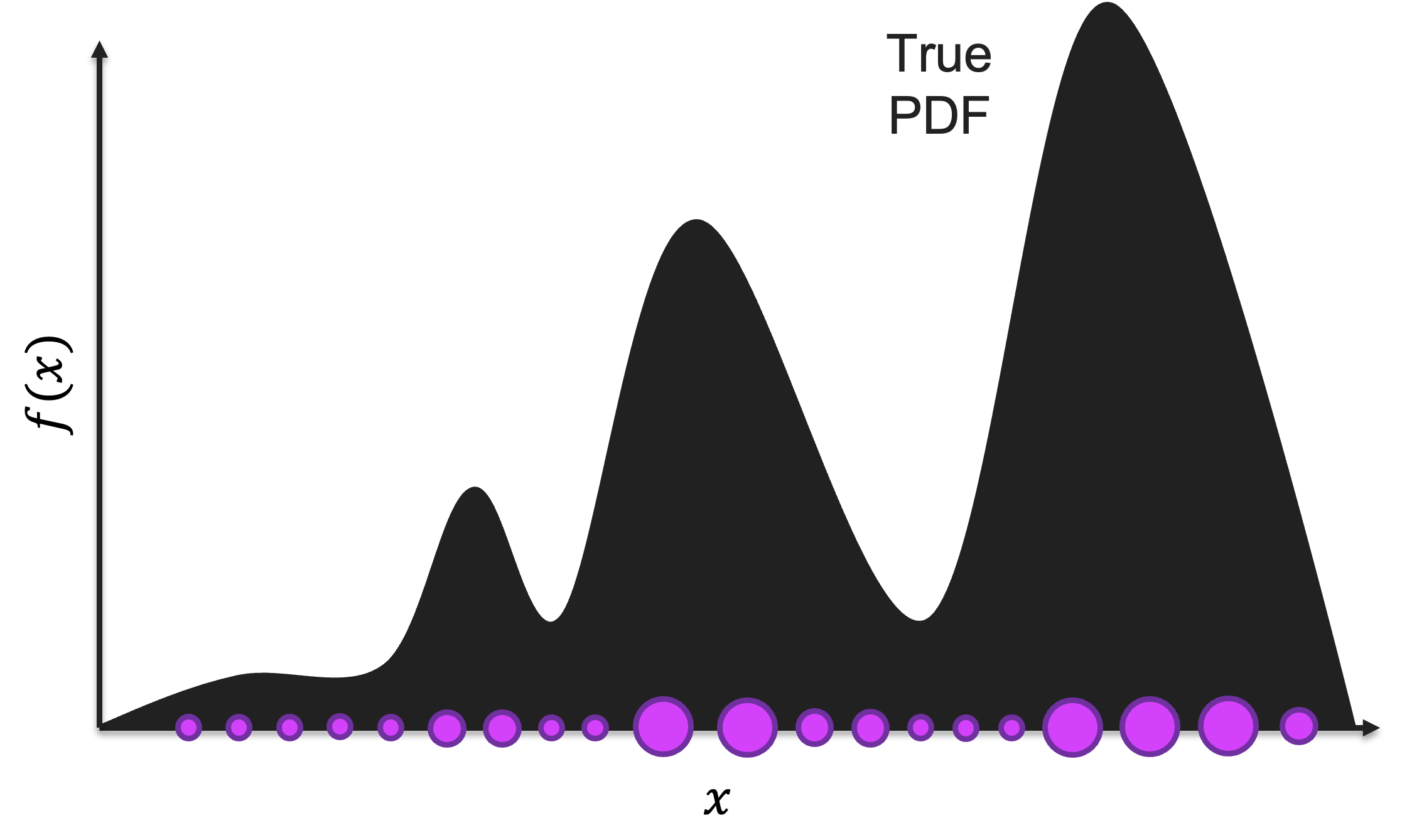}} \hfill
    \subfloat[Approximation of PGM filters.]{\includegraphics[clip,width=0.33\linewidth]{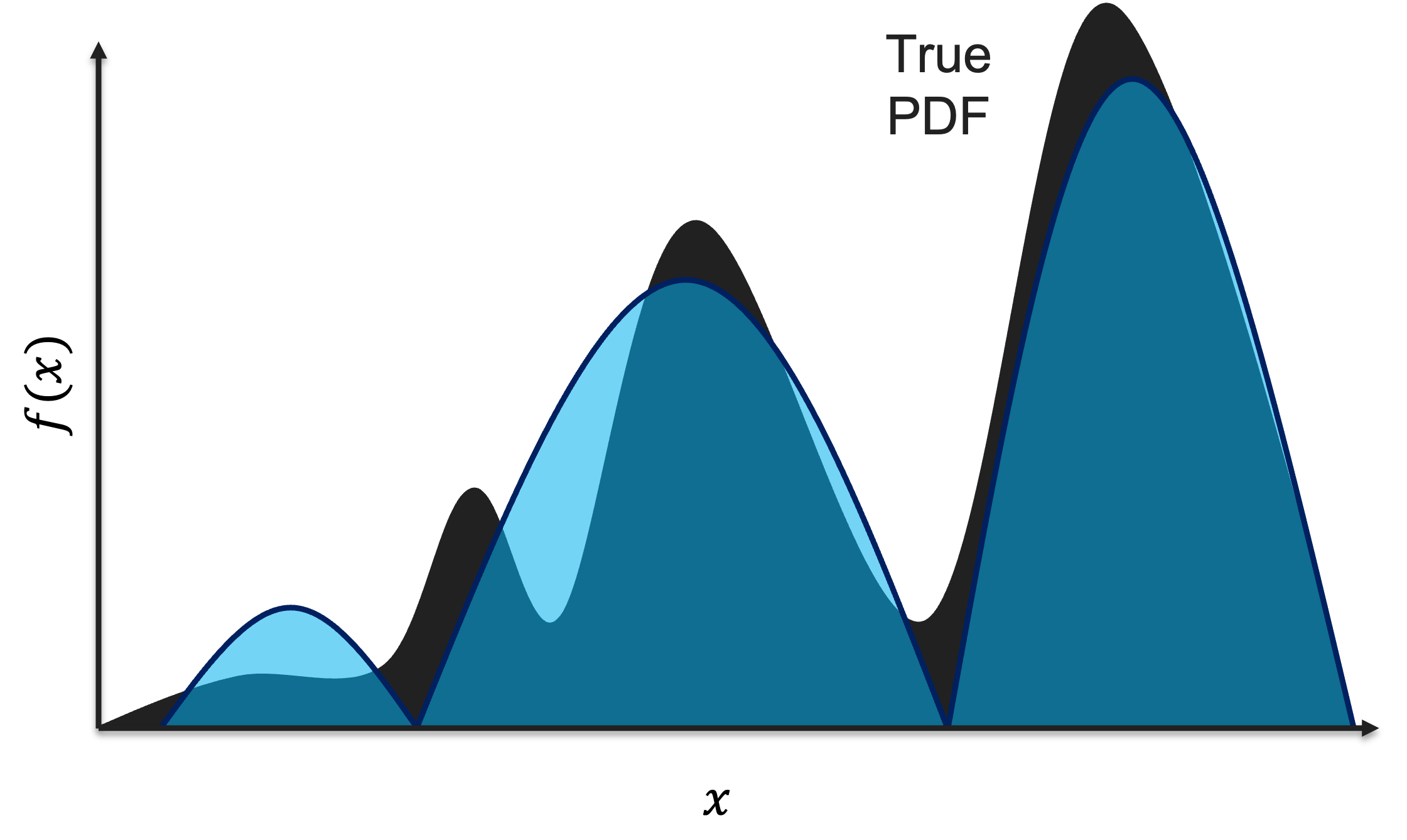}} \hfill
    \subfloat[Approximation of proposed filter.]{\includegraphics[clip,width=0.33\linewidth]{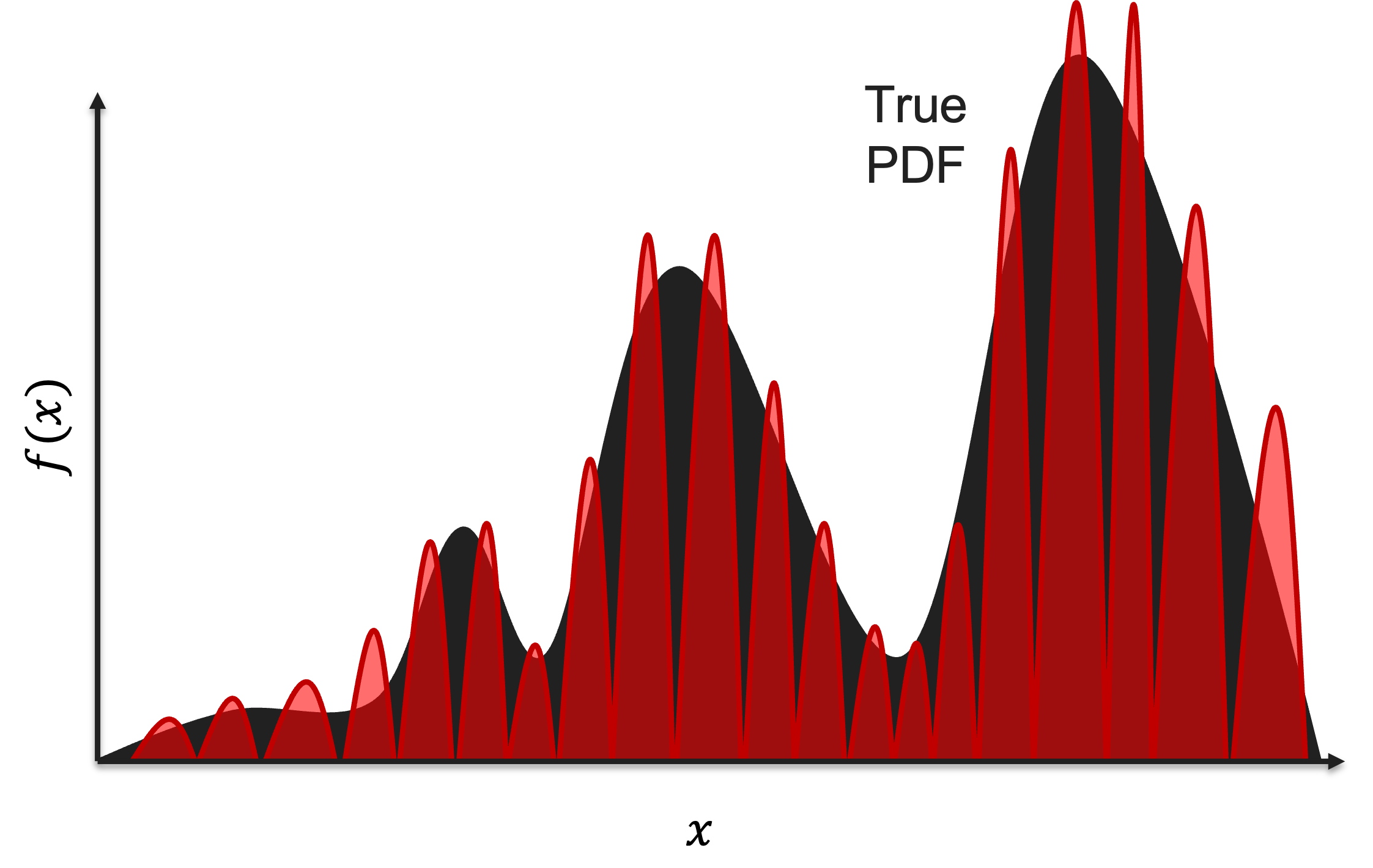}}
\caption{\textbf{A graphical conceptualization of the PDF approximation by different approaches.} PF, PGM filters, and the proposed filter. PFs use particles and the weight of each particle, PGM filters cluster the particles and construct the GMM PDF using clusters, and GMF considers every sample as a Gaussian distribution.}
\label{fig:comparison}
\end{figure*}

A comparison of PF, PGM filters and the proposed GMF filter approaches is shown in Table \ref{tab:comparison} and the graphical illustration of different ways of PDF approximation of these approaches are shown in Fig. \ref{fig:comparison}.

\subsection{Initialization}
The filter is initialized with $N_p$ samples drawn from the Gaussian distribution and the covariance of each Gaussian component is initialized as follows:

\begin{equation}
  \bm{x}_0^i \sim \mathcal{N}(\bm{\mu}_0, \bm{P}_0), \quad i = 1, \ldots, N_p, \label{eq4}
\end{equation}

\begin{equation}
  \bm{P}_0^i = h^2 \bm{I}_{n \times n}, \label{eq5}
\end{equation}

\noindent where $h = N_p^{-0.2}$ is the bandwidth parameter with the exponent $-0.2$ (or $-1/5$) that is a commonly used rule of thumb to balance the trade-off between bias and variance in kernel density estimation \cite{silverman_h, stordal_h}. This bandwidth parameter determines the relative width of the covariance based on the number of samples, $N_p$, to minimize the overlap between each Gaussian component. 

\subsection{Time Propagation}
Mean and covariance estimates in Eqs. (\ref{eq4}) and (\ref{eq5}) of each Gaussian component are propagated independently with dynamics $f(\cdot)$ and the Jacobian of dynamics in given time step $k$, $\bm{F}_{k}^i$ as follows:

\begin{equation}
  \bm{x}_{k|k-1}^i = f(\bm{x}_{k-1}^i, \bm{w}_k), \label{eq6}
\end{equation}

\begin{equation}
  \bm{P}_{k|k-1}^i = \bm{F}_{k}^i\bm{P}_{k-1}^i{\bm{F}_{k}^i}^T + \bm{\Gamma}_k \bm{Q}_k \bm{\Gamma}_k^T, \label{eq7}
\end{equation}

\noindent where $\bm{w}_k \sim \mathcal{N}(0, \bm{Q}_k)$ is a zero-mean Gaussian process noise with covariance $\bm{Q}_k$ and $\bm{\Gamma}_k$ is a process noise model. To prevent the overlap between Gaussian components from becoming too large, we adopt the idea of substituting the covariance if the covariance is bigger than the upper limit of covariance with scaled sample covariance as follows:

\begin{equation}
  \bm{P}_{k|k-1}^{\text{scaled}} = h^2 \frac{1}{N_p-1} \sum_{i=1}^{N_p} (\bm{x}_{k|k-1}^i - \bar{\bm{x}}_k)(\bm{x}_{k|k-1}^i - \bar{\bm{x}}_k)^T, \label{eq8}
\end{equation}

\noindent where the sample covariance of the samples is scaled down with bandwidth parameter $h$ to ensure that the covariance does not become overly large while calculating the covariance adaptively. It is worth noting that, in this study, the scaled sample covariance is introduced to replace the the predefined covariance in the MI bounding step which is the distinct point from the original idea in \cite{Psiaki_GaussianSum, Psiaki_Blob}.

\subsection{Measurement Update}
Each Gaussian component with time propagated mean and covariance (or scaled sample covariance) is updated with measurement from the sensor and the process of measurement update is the same as an EKF. The additional calculation of the weight $w^i$ of each Gaussian component, using the Gaussian likelihood update (as in PF) is required to reconstruct the GMM. These processes are shown as follows:

\begin{equation}
  \bm{S}_k^i = \bm{H}_{k}^i \bm{P}_{k|k-1}^i {\bm{H}_k^i}^T + \bm{R}_{k}, \label{eq9}
\end{equation}

\begin{equation}
  \bm{K}_k^i = \bm{P}_{k|k-1}^i {\bm{H}_k^i}^T (\bm{S}_k^i)^{-1}, \label{eq10}
\end{equation}

\begin{equation}
  w_k^i = w_{k-1}^i \cdot \frac{e^{-(\bm{z}_k - \bm{H}_k^i\bm{x}_{k|k-1}^i)^2 / 2\sigma}}{\sqrt{2 \pi \bm{R}_k}}, \label{eq11}
\end{equation}

\begin{equation}
  \bm{x}_{k}^i = \bm{x}_{k|k-1}^i + \bm{K}_k^i(\bm{z}_k - \bm{H}_k^i\bm{x}_{k|k-1}^i), \label{eq12}
\end{equation}

\begin{equation}
  \bm{P}_{k}^i = (\bm{I} - \bm{K}_k^i \bm{H}_k^i)\bm{P}_{k|k-1}^i, \label{eq13}
\end{equation}

\noindent where $\bm{S}_k^i$ is the innovation covariance with measurement noise covariance $\bm{R}_k$, $\bm{K}_k^i$ is the Kalman gain for the $i^{th}$ sample and $\bm{z}_k$ is the measurement at $k^{th}$ time step.

\subsection{Resampling}
The samples are then resampled from the GMM after the measurement update as follows:

\begin{equation}
  \bm{x}_{k}^i \sim \sum_{i=1}^{N_p} w_k^i \mathcal{N}(\bm{x}_{k}^i, \bm{P}_{k}^i). \label{eq14}
\end{equation}

Also, the covariance of each resampled sample is reset with the scaled sample covariance of Eq. \ref{eq8}. This step prevents the sample degeneracy issue and ensures the covariance of each Gaussian component is small enough to represent the posterior distribution precisely.

\section{Numerical Simulation Results and Analysis: Blind Tricyclist Problem}
\label{Numerical}
The blind tricyclist problem~\cite{Psiaki_Tricyclist} is a benchmark example for the nonlinear filters proposed by Psiaki. This problem considers a tricyclist navigating around a park based only on measurements of the two relative bearings where these measurements are intermittent. This problem is highly nonlinear with low observability. The 7-dimensional state vector, 

\begin{equation}
    \bm{x} = [X, Y, \theta, \phi_1, \dot{\phi}_1, \phi_2, \dot{\phi}_2]^T,
    \label{eq15}
\end{equation}

\noindent represents the position and heading of the tricyclist, as well as the initial angles and angular velocities of two friends on a merry-go-round. The 2-dimensional measurement vector, 

\begin{equation}
    \bm{y} = [\psi_1, \psi_2]^T, \label{eq16}
\end{equation}

\noindent consists of the relative bearings between the tricyclist and the two friends, respectively. This example was adopted for many nonlinear filters and is used here to validate the performance of the proposed filter and the implementations of the compared filters. Further details of the problem can be found in \cite{Psiaki_Tricyclist}. 

The proposed filter is compared with the state-of-the-practice PF and PGM filters, as other filters were evaluated in previous studies. All filters were utilized with 7,000 samples, the process noise covariance $Q = \text{diag}([0.0567, 0.0000, 0.0063, 0.0063, 0.0000])$ and the measurement noise covariance $R = 10^{-3} \begin{bsmallmatrix} 0.3046 & 0 \\ 0 & 0.1354 \end{bsmallmatrix}.$ The initial state for each Monte Carlo (MC) simulation is selected with an initial covariance matrix $P_0 = \text{diag}([18.75^2, 18.75^2, (5\pi/8)^2, (5\pi/6)^2, (5\pi/6)^2, (1.857 \times 10^{-2})^2, (1.857 \times 10^{-2})^2])$. The bandwidth parameter $h$ of the proposed filter, the minimum points $minPts$ and the radius $\epsilon$ used for clustering in DBSCAN for PGM filters, and the UT parameters for PGM-DU filter are shown in Table \ref{tab:parameter_blind}. \looseness=-1

\begin{table}[h!]
\centering
\caption{Parameters of the tested filters for the blind tricyclist benchmark problem.}
\begin{tabular}{ l l l l }
\toprule
\textbf{Algorithm} & \textbf{Module} & \textbf{Parameter} & \textbf{Value} \\
\midrule
Proposed & - & $h$ & $N_p^{-0.2}$ \\
\multirow{2}{*}{PGM Filters} & \multirow{2}{*}{DBSCAN} & $minPts$ & 8 \\
&  & $\epsilon$ & 5.0 \\
\multirow{3}{*}{PGM-DU} & \multirow{3}{*}{UT} & $\alpha$ & 0.01 \\
&   & $\beta$ & 2 \\
&   & $\kappa$ & 0 \\
\bottomrule
\end{tabular}
\label{tab:parameter_blind}
\end{table}

In order to compare the performance of the proposed filter with other filters, Root Mean Square Error (RMSE) of terminated position, average RMSE throughout the simulation time (ARMSE) and average simulation time are considered as performance metrics. Average simulation time here refers to the required time to run one MC simulation. A total number of 100 MC simulations is conducted and the results are shown in Table \ref{tab:results_blind}. \looseness=-1

\begin{table}[h!]
\centering
\caption{Results of blind tricyclist problem.}
\begin{tabular}{ l l l l }
\toprule
\textbf{Algorithm} & \textbf{\makecell[l]{Terminate \\ RMSE (m)}} & \textbf{\makecell[l]{Position \\ ARMSE (m)}} & \textbf{\makecell[l]{Average simulation \\ time (sec)}} \\
\midrule
\cellcolor{yellow!25}\textbf{Proposed} & \cellcolor{yellow!25}\textbf{5.6348} & \cellcolor{yellow!25}\textbf{13.8535} & \cellcolor{yellow!25}113.0153 \\
PGM-DS & 119.6745 & 101.8103 & 236.0642 \\
PGM-DU & 10.2320 & 22.6147 & 254.1003 \\
PF & 11.9186 & 26.1625 & \textbf{49.5318} \\
\bottomrule
\end{tabular}
\label{tab:results_blind}
\end{table}

\begin{figure}[thpb]
  \centering
  \includegraphics[scale=0.11]{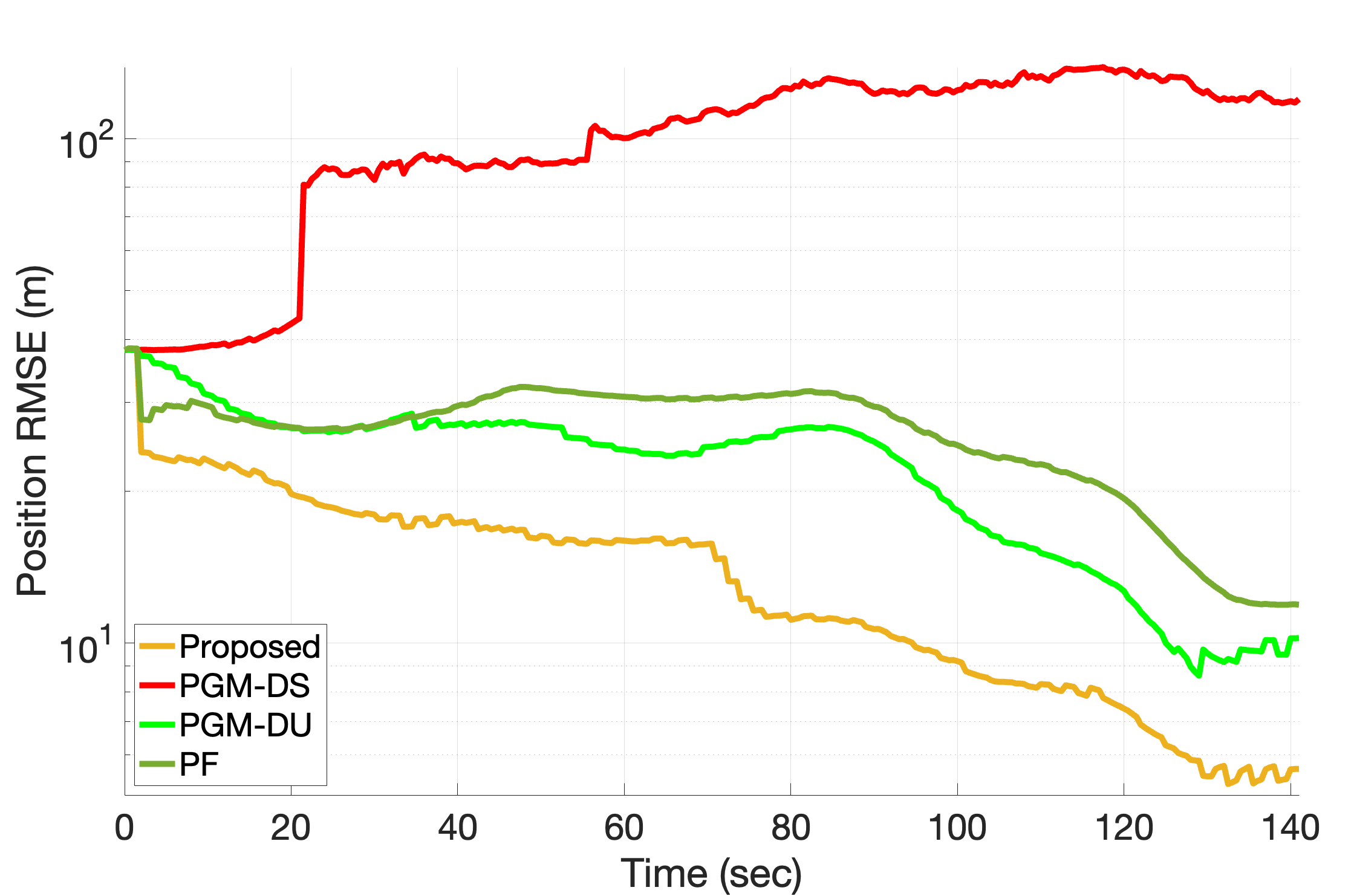}
  \caption{\textbf{Results of blind tricyclist problem.} The proposed filter showed the faster convergence over other compared filter with smaller terminate RMSE and ARMSE.}
  \label{fig:blind_rmse}
\end{figure}

The results show that the proposed filter performs better than compared filters in terms of estimation error: terminate RMSE and ARMSE. Also, the proposed filter can reduce the estimation time by half compared to PGM filters. The convergence of RMSE of the simulation example is shown in Fig. \ref{fig:blind_rmse}. \looseness=-1

\section{Experimental Results for an Autonomous Robotic Radio Signal Source Search and Localization Mission}
\label{Scenario}
\subsection{Overview}
\begin{figure}[thpb]
  \centering
  \includegraphics[scale=0.27]{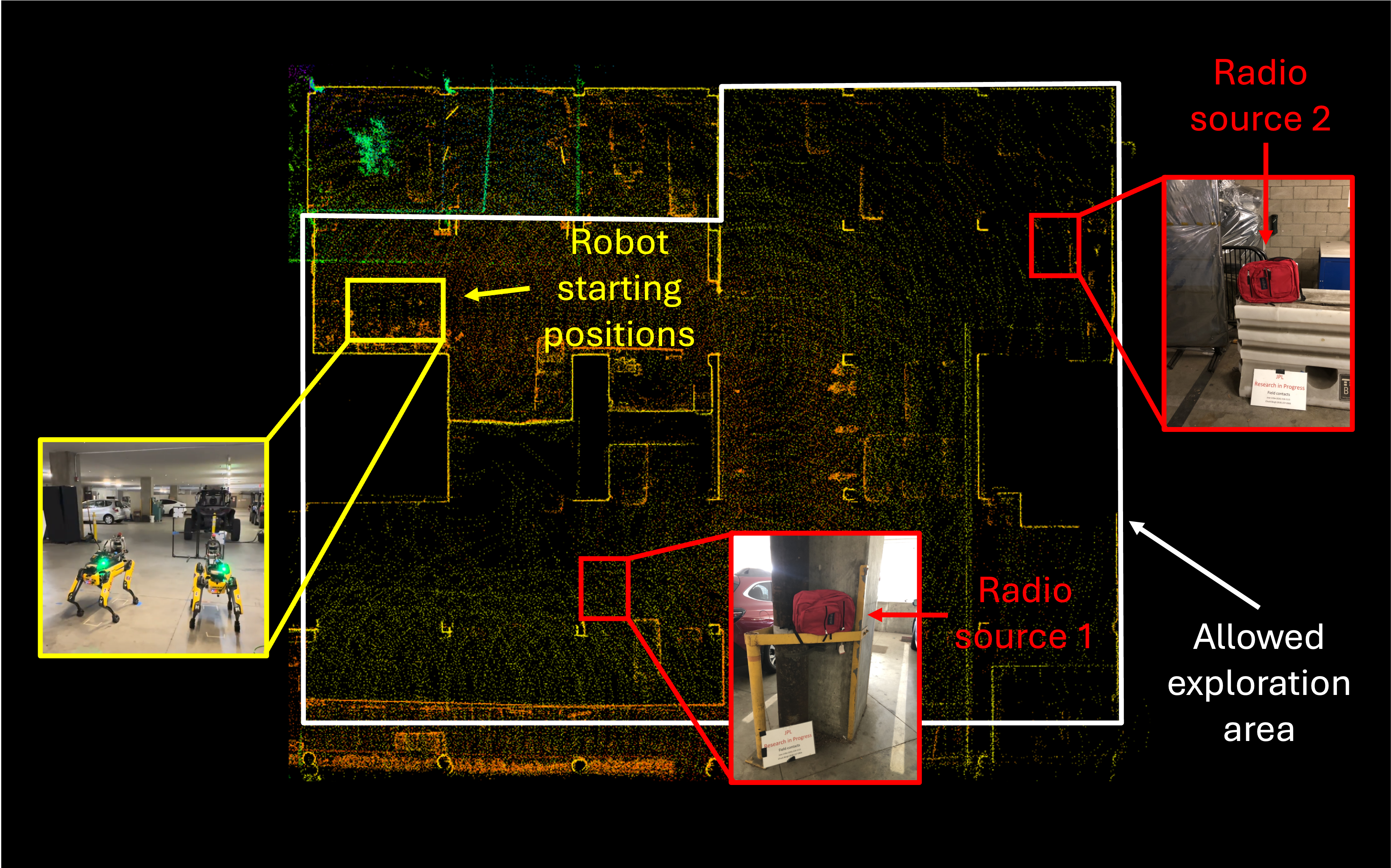}
  \caption{\textbf{Example test environment.} The yellow box indicates the starting positions of the robots, while the two red boxes mark the locations of the radio signal sources. The white box represents the allowed exploration area.}
  \label{fig:test_environment}
\end{figure}

For demonstration towards real-world applications, the scenario for an autonomous multi-robot search and localization of multiple radio signal sources in an unknown environment is considered. In this study, we considered scenarios in which two legged robots (Boston Dynamics Spot), equipped with custom compute and sensing payloads~\cite{Bouman_Spot}, attempt to localize two radio signal source simultaneously but independently. Each robot autonomously explores the unknown environment and localizes its tasked radio signal source independently, without utilizing any prior information. Fig. \ref{fig:test_environment} shows the experiment setup presentation the mission environment map (shown for illustration here), the allowed exploration area marked with white boundary, the starting location of the two robots marked in yellow, and the unknown locations of the the two radio sources marked in red. 

This study is focused on the estimation of the position of the radio signal sources and the exploration methodology and Simultaneously Localization and Mapping (SLAM) techniques fall outside of this study's scope. Further detail for autonomous exploration~\cite{exploration_nebula} and SLAM~\cite{slam_rose} are available in previous studies. \looseness=-1

\subsection{Measurement Model}\label{SM}
Radios operating in the 5 GHz range are utilized as a signal sources to localize. Two radios are placed in an unknown space with their signal being measured by the two robots. Only the Signal-to-Noise Ratio (SNR) value is measured by the robots as an indicator radio signal strength with respect to distance from the robots. In other words, no direct manner of measuring the distance or bearing between the robot and the signal source. Furthermore, the SNR value is a noisy indicator of radio distance from the robot as LOS and Non-LOS (NLOS) SNR measurements can result in the same value while at different distances from the robot. A bimodal log-linear measurement model is considered to predict signal strength given the distance between the position of the robot and the estimated sample radio location, $d_i$. The predicted signal strength for the $i^{th}$ sample, $\bm{z}_k^i$ can be calculated with two tuning parameters, log-linear multiplier $p_{1}$ and log-linear constant $p_{2}$ as follows:

\begin{equation}
  \bm{z}_k^i = p_{1} log_{10}(d_i)+p_{2}. \label{eq17}
\end{equation}

The term ``bimodal'' here refers to the consideration of the probability that the signal source being on either the LOS and NLOS at each measurement step. The predicted signal strength in Eq. \ref{eq17} is calculated twice: once for the LOS case and once for the NLOS case, each with different parameters $p_{1}$ and $p_{2}$. To determine the probability of being on LOS or NLOS, the filter first check if the sample is within a threshold distance from the position of the robot. Then, if the sample is within the threshold distance, the probability of the sample being on the LOS is set to 0.95, while on the NLOS is set to 0.05. On the other hand, if the sample is not within the threshold distance, the probability of the sample being LOS is set to 0.05 and on the NLOS to 0.5. Finally, the weight of each sample is computed as a linear combination of the LOS and NLOS cases, weighted by their respective probabilities. It is worth noting that given a measurement that can be interpreted as a range-only without bearing information, and the robot's current position, the belief of the state can be ring-shaped.

\subsection{Evaluation Metrics}\label{EM}
In order to evaluate and compare the tested filters, two evaluation metrics are considered. The average position error estimation is the first metric to evaluate the estimation accuracy of the filters throughout the test time. It is calculated by time-averaging the distance between the estimated mean and the true position of the radio signal source. This metric is chosen to show the accuracy of the proposed and compared filters. The second metric is the average iteration time, which is used to evaluate the computation cost of the presented methods. This metric is important to ensure that the proposed and compared methods are suitable for real-time online applications towards real-world robotic applications. The average iteration time is calculated by averaging the time elapsed during each iteration of the filter throughout the experiment. In addition, the average number of Gaussian mixtures is presented for a monitoring purpose. This is to monitor how many clusters are considered to construct the GMM for the proposed filter and PGM filters. Note that the proposed filter considers all samples as individual Gaussian distributions and the number of mixtures remains constant. \looseness=-1

\subsection{Experimental Results and Discussion}
The proposed filter was tested with real-world data of the considered scenarios as described in Section \ref{Scenario} and compared against the performance of two PGM filters, PGM-DS and PGM-DU filters and PF. All filters were utilized with 3,000 samples, the process noise covariance $Q = 0.5$ and the measurement noise covariance as a linear combination of $R_{min} = 8$ and $R_{max} = 35$ based on the SNR. This adaptive measurement noise covariance allows the filter to adjust dynamically based on the SNR, making it trust the measurements more as it gets closer to the source with a higher SNR. The bandwidth parameter $h$ of the proposed filter, the minimum points $minPts$ and the radius $\epsilon$ used used for clustering in DBSCAN for PGM filters, and the UT parameters for PGM-DU filter are shown in Table \ref{tab:parameter}.

\begin{table}[h!]
\centering
\caption{Parameters of the tested filters for the autonomous robotic radio signal source search and localization mission.}
\begin{tabular}{ l l l l }
\toprule
\textbf{Algorithm} & \textbf{Module} & \textbf{Parameter} & \textbf{Value} \\
\midrule
Proposed & - & $h$ & $N_p^{-0.2}$ \\
\multirow{2}{*}{PGM Filters} & \multirow{2}{*}{DBSCAN} & $minPts$ & 4 \\
&  & $\epsilon$ & 1.0 \\
\multirow{3}{*}{PGM-DU} & \multirow{3}{*}{UT} & $\alpha$ & 0.001 \\
&   & $\beta$ & 2 \\
&   & $\kappa$ & 0 \\
\bottomrule
\end{tabular}
\label{tab:parameter}
\end{table}

\begin{figure*}[t!]
\centering
\subfloat[Test results of scenario 1.]{
       \includegraphics[clip,width=0.47\linewidth]{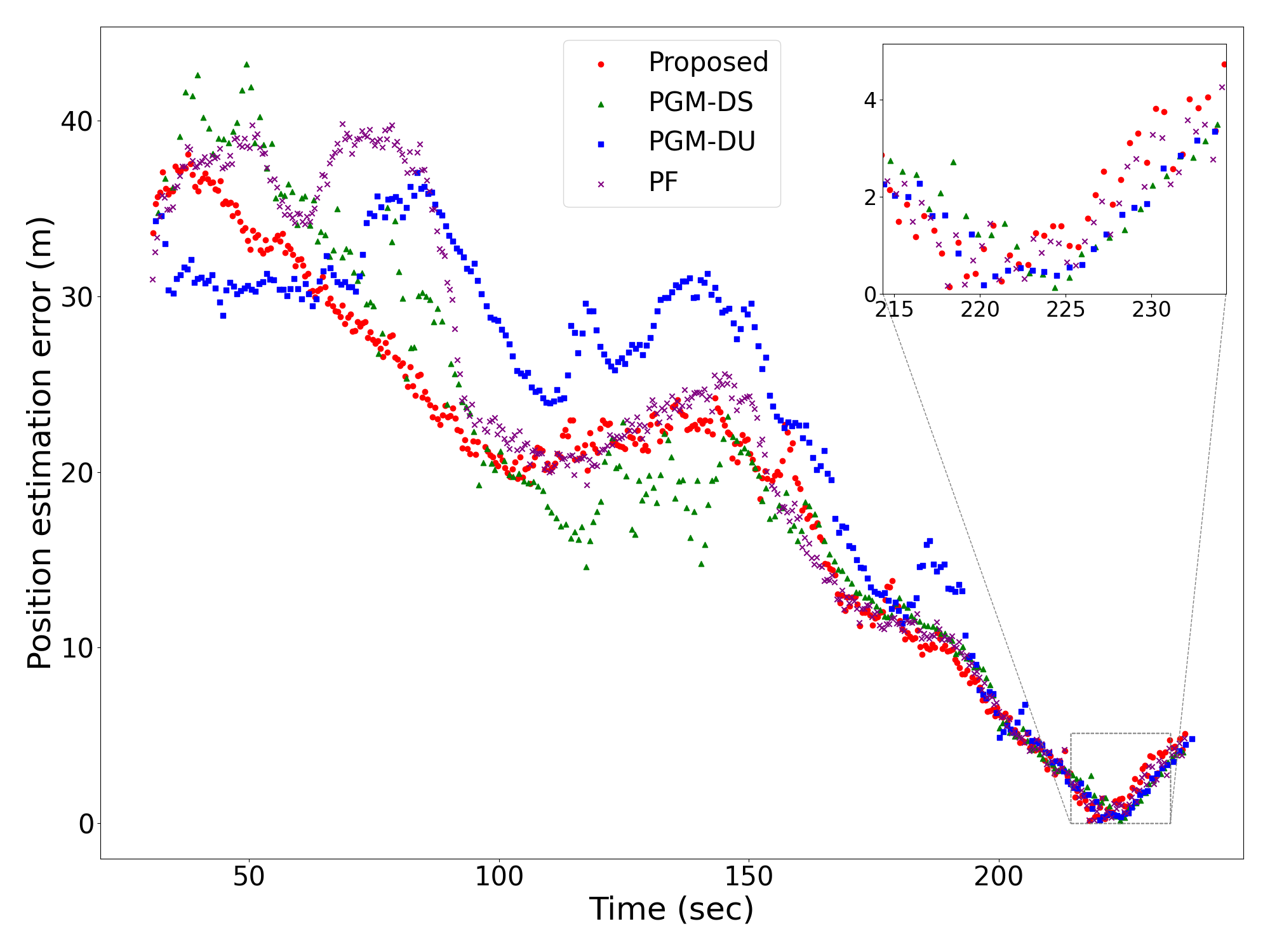}} \,
\subfloat[Violin plot of scenario 1.]{
       \includegraphics[clip,width=0.47\linewidth]{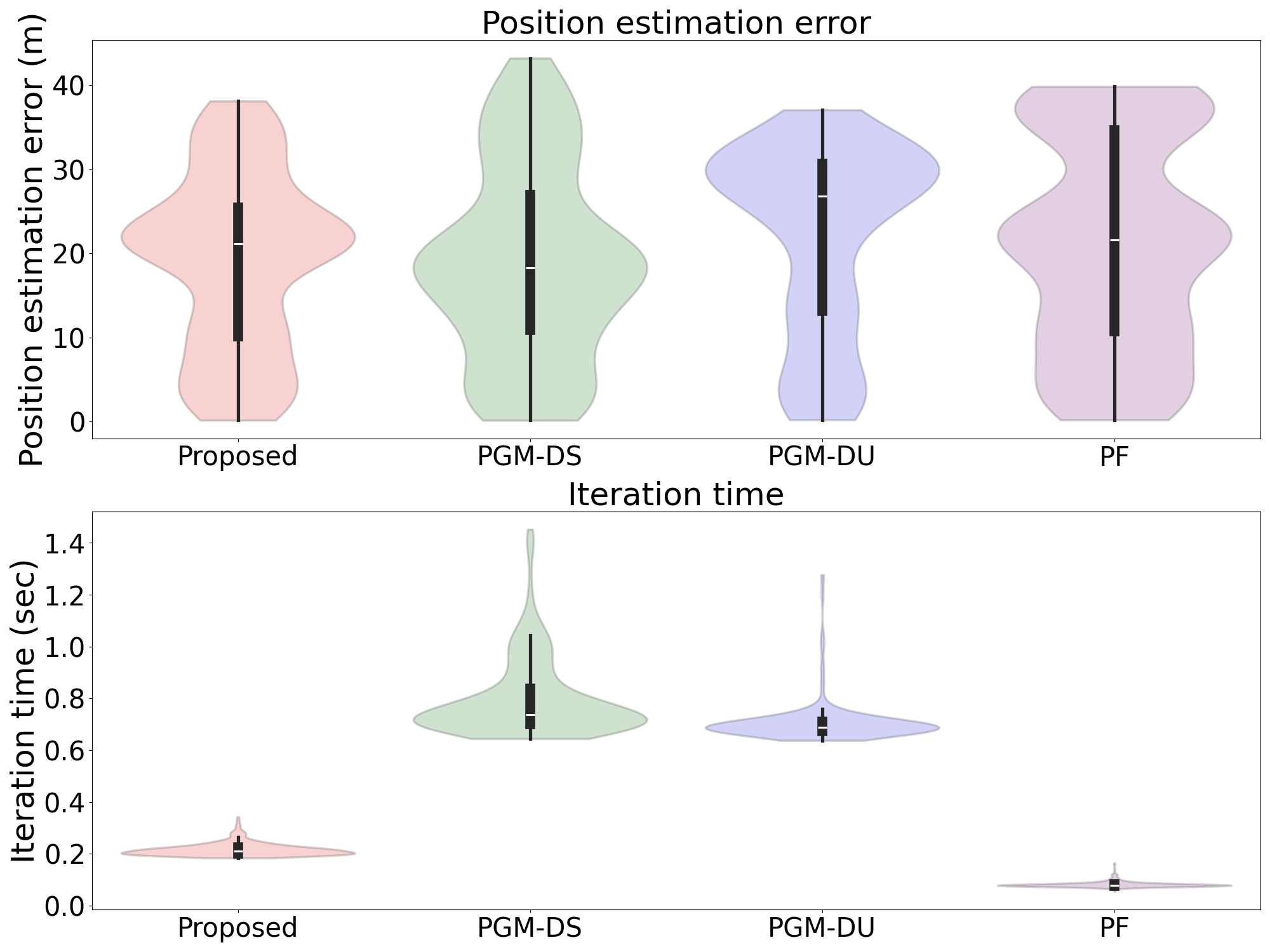}} \, 
       \subfloat[Test results of scenario 2.]{
       \includegraphics[clip,width=0.47\linewidth]{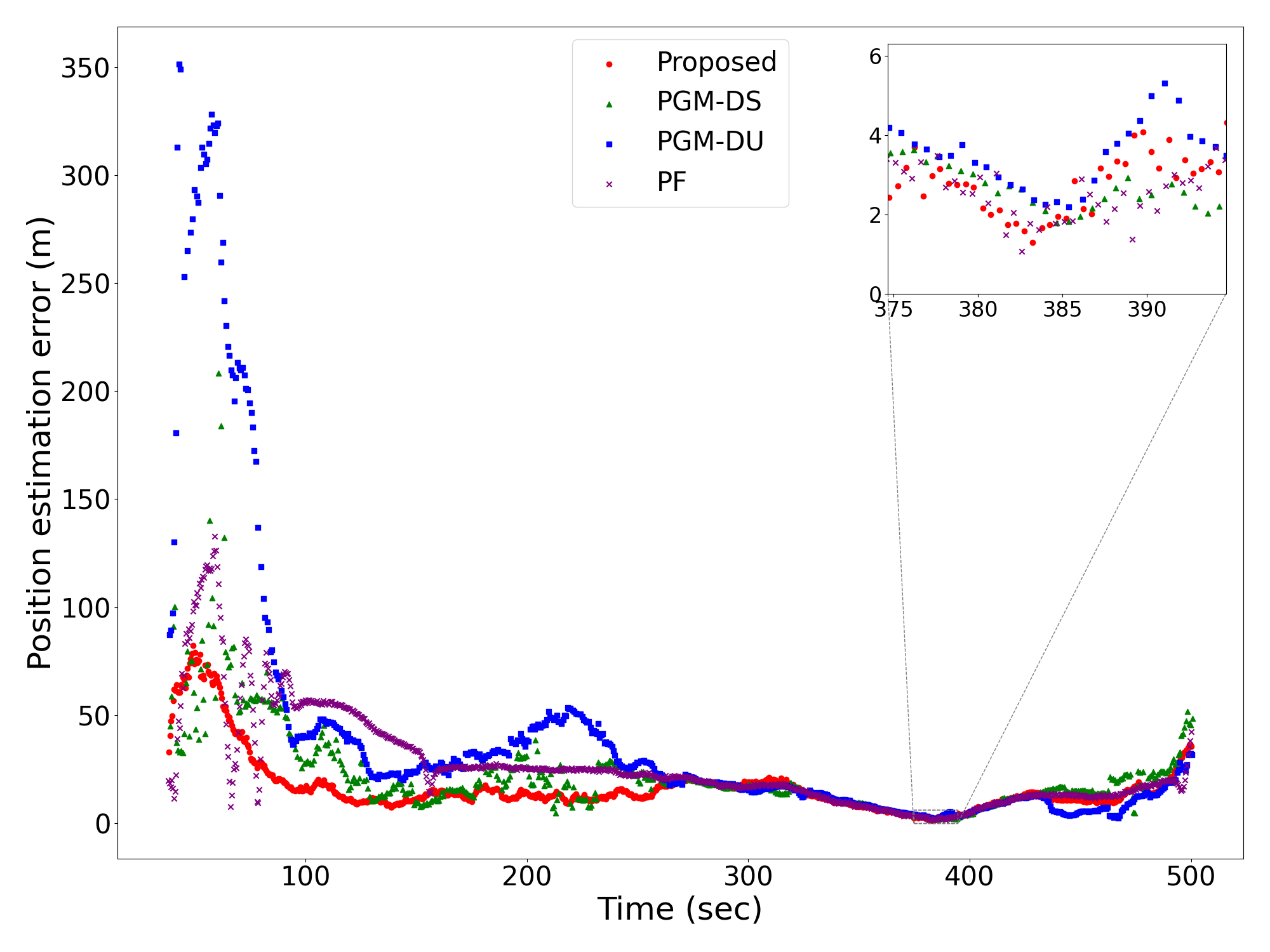}} \,
\subfloat[Violin plot of scenario 2.]{
       \includegraphics[clip,width=0.47\linewidth]{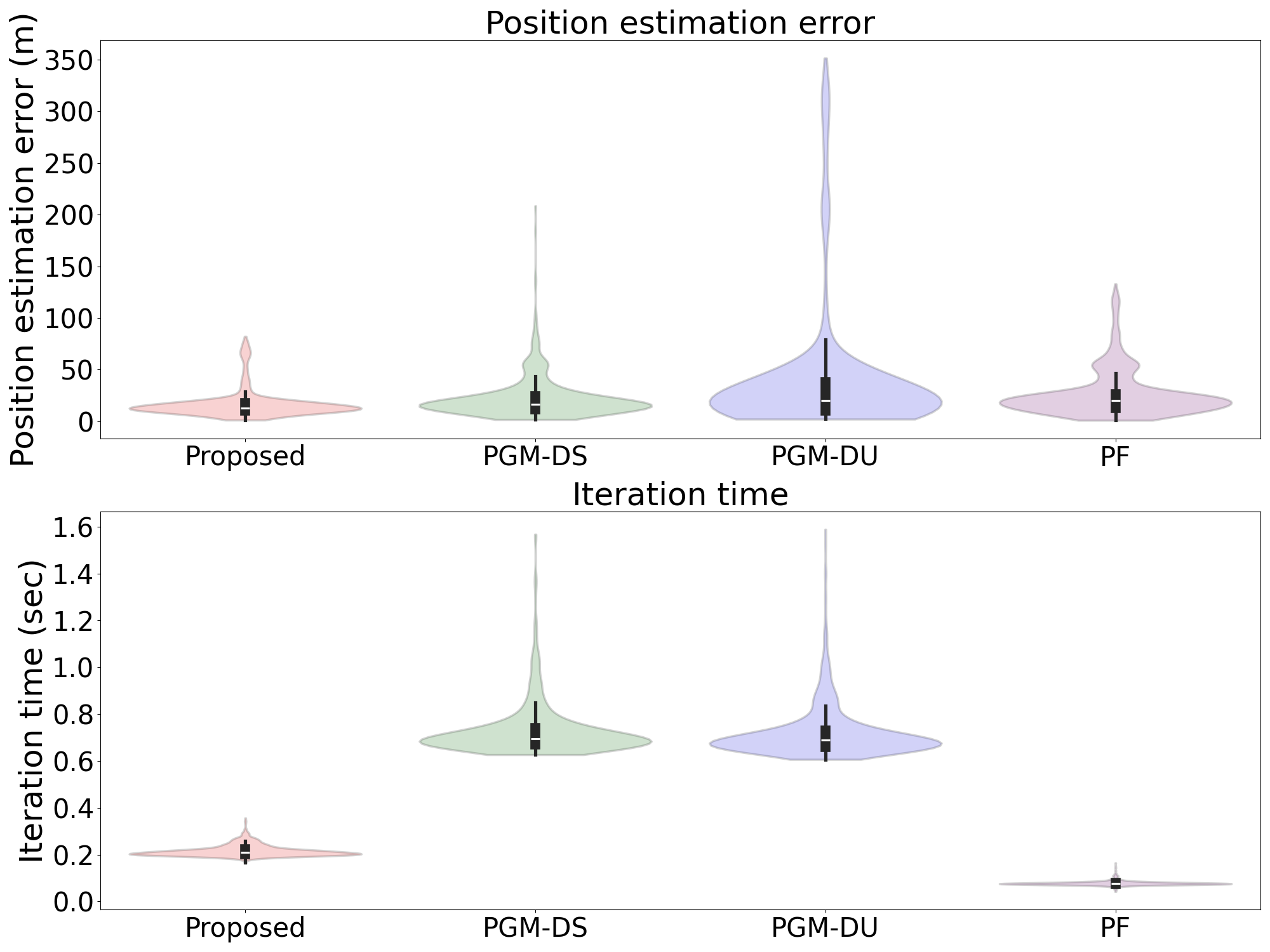}} 
\caption{Evaluation of the proposed approach against benchmarks (PGM-DS, PGM-DU, and PF) regarding i) position estimation error profiles over operational time and ii) statistical distributions in position estimation errors and iteration times. Note that sub-figures (a) and (c) correspond to Scenario 1 and other sub-figures (b) and (d) related to Scenario 2.}
\label{fig:exp_rosbag_results}
\end{figure*}

Two experiments were conducted to evaluate the performance of all methods. In the first experiment, the radio signal sources were placed in the direct LOS of the robot to mitigate any effects of radio signal reflection from the environment. In the second experiment, the radio signal sources were placed outside the direct LOS of the robot and hidden closer corners in the unknown environment. Each experimental scenario contained two radio signal sources to be searched for and localized by the two robots. Note that each robot localizes each target independently as already mentioned in Section~\ref{Scenario}. The test results of the search and localization and standard deviations for the two experimental scenarios are presented in Tables \ref{tab:scn1} and \ref{tab:scn2}, as well as Fig. \ref{fig:exp_rosbag_results}.

\begin{table}[h!]
\centering
\caption{Results of scenario 1.}
\begin{tabular}{ l l l l }
\toprule
\textbf{Algorithm} & \textbf{\makecell[l]{Average \\ position \\ estimation \\ error (m)}} & \textbf{\makecell[l]{Average \\ iteration \\ time (sec)}} & \textbf{\makecell[l]{Average \\ number of \\ mixtures}} \\
\midrule
\cellcolor{yellow!25}\textbf{Proposed} & \cellcolor{yellow!25}19.0 & \cellcolor{yellow!25}0.22 & \cellcolor{yellow!25}3000 \\
PGM-DS & \textbf{18.8} & 0.80 & 30.6 \\
PGM-DU & 21.9 & 0.71 & 4.9 \\
PF & 20.9 & \textbf{0.08} & - \\
\bottomrule
\end{tabular}
\label{tab:scn1}
\end{table}

\begin{table}[h!]
\centering
\caption{Results of scenario 2.}
\begin{tabular}{ l l l l }
\toprule
\textbf{Algorithm} & \textbf{\makecell[l]{Average \\ position \\ estimation \\ error (m)}} & \textbf{\makecell[l]{Average \\ iteration \\ time (sec)}} & \textbf{\makecell[l]{Average \\ number of \\ mixtures}} \\
\midrule
\cellcolor{yellow!25}\textbf{Proposed} & \cellcolor{yellow!25}\textbf{16.9} & \cellcolor{yellow!25}0.22 & \cellcolor{yellow!25}3000 \\
PGM-DS & 22.4 & 0.73 & 19.0 \\
PGM-DU & 40.8 & 0.72 & 21.5 \\
PF & 26.7 & \textbf{0.08} & - \\
\bottomrule
\end{tabular}
\label{tab:scn2}
\end{table}

\begin{figure*}[t!]
\centering
    \begin{NiceTabular}{ccc}
        \Block[borders={right,tikz={densely dashed, thick}}]{1-1}{}
        \subfloat[Averaged comparison from both scenarios.]{\includegraphics[clip,width=0.315\linewidth]{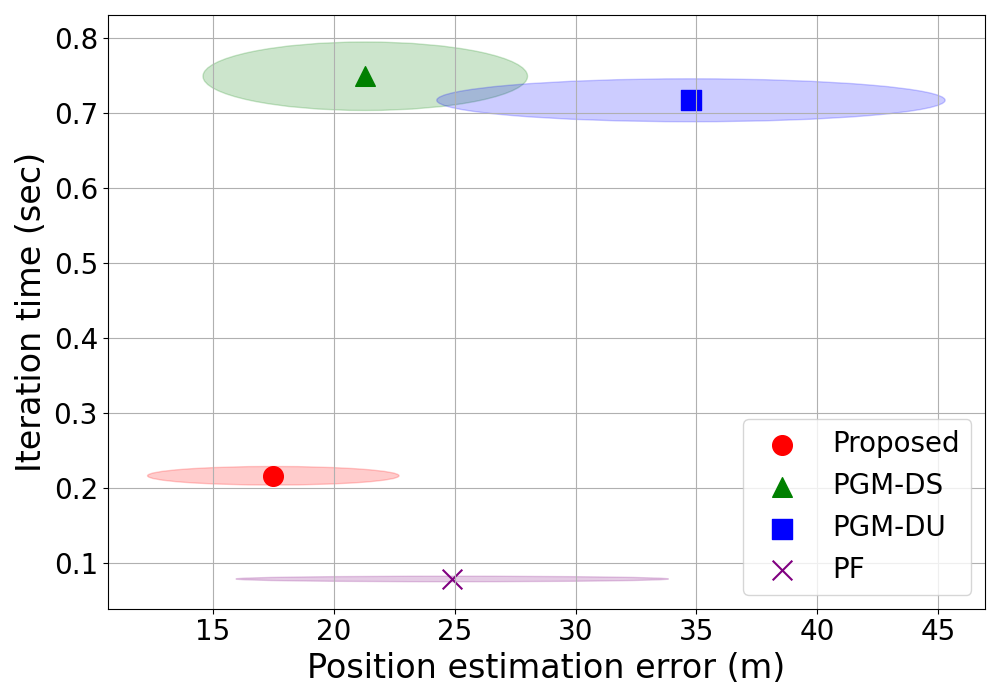}} &
        \subfloat[Performance comparison in scenario 1.]{\includegraphics[clip,width=0.315\linewidth]{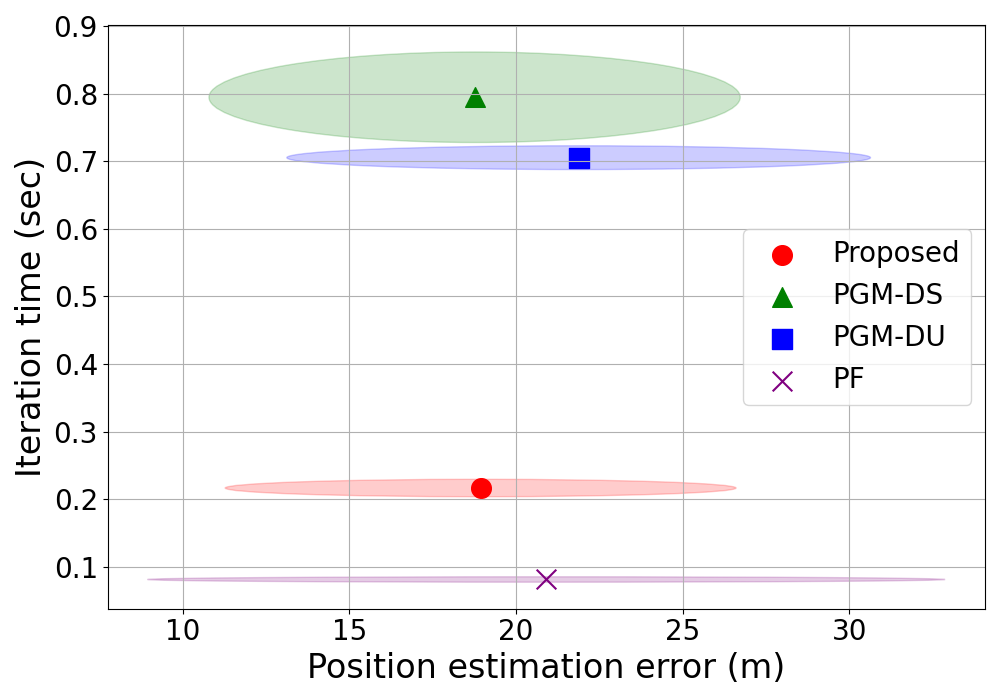}} &
        \subfloat[Performance comparison in scenario 2.]{\includegraphics[clip,width=0.315\linewidth]{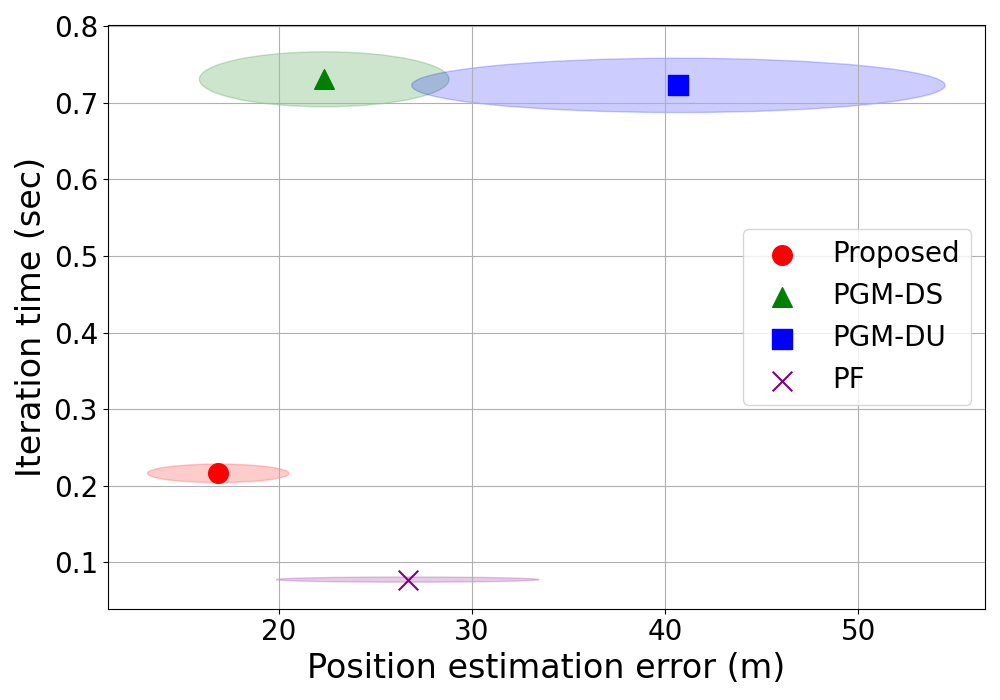}}
    \end{NiceTabular}
\caption{The visualized comparison in terms of position estimation error (x-axis) and iteration time (y-axis). (a) shows the averaged results from both two scenarios, whilst (b) and (c) are results from Scenario 1 and Scenario 2 respectively. Note that markers and ovals represent mean and region within 2-$\sigma$ bounds in distribution respectively.}
\label{fig:exp_matric_analysis}
\end{figure*}

In the first scenario, the proposed filter showed an average position estimation error of 19.0~m whilst the other filters, PGM-DS, PGM-DU and PF showed 18.8~m, 21.9~m and 20.9~m, respectively. Moreover, the average iteration time of the proposed filter was 0.22~seconds whilst the other filters were 0.80~seconds, 0.71~seconds and 0.08~seconds, respectively. In the second scenario, the proposed filter showed an average position estimation error of 16.9~m whilst the other filters, PGM-DS, PGM-DU and PF showed 22.4~m, 40.8~m and 26.7~m, respectively. Moreover, the average iteration time of the proposed filter was 0.22~seconds whilst the other filters were 0.73~seconds, 0.72~seconds and 0.08~seconds, respectively. It is worth noting that the proposed filter and PF were able to estimate the belief with a ring shape, whereas the PGM filters were not able to do so, as expected. A summary of the performance of each filter on average from both scenarios and the tested scenarios is shown in Fig. \ref{fig:exp_matric_analysis}. \looseness=-1

The test results of the first scenario, where the radio signal sources were located on the LOS of the robots, showed that the proposed filter performs the second-best in average position estimation error of 19.0~m whereas PGM-DS showed the smallest error of 18.8~m. Although the average position error of the proposed filter was larger than PGM-DS, the proposed filter was 3.64 times faster than PGM-DS. In the second scenario, where the radio signal sources were located outside the LOS of the robots, the proposed filter showed the smallest average position estimation error of 16.9~m which is 24.6$\%$ improvement from the second-best performance, 22.4~m of PGM-DS. In this case, the proposed filter was again 3.32 times faster than PGM-DS. Throughout the scenarios, PF was the fastest and the proposed filter was the second fastest whilst PGM-DS and PGM-DU were approximately three times slower than the proposed filter. It is worth mentioning that the average iteration time of each tested filter was less than one second. This indicates that all filters can be utilized in online estimation applications with a 1~Hz update, which is the case of the tested environment.~\looseness=-1

In addition to the improved performance with less computation than PGM filters, the proposed filter showed smaller deviation in both position error and iteration time as is shown in Fig. \ref{fig:exp_rosbag_results} (b) and (d). In the first scenario, PGM-DU and PF showed high occurrence in the high error region whilst the proposed filter and PGM-DS filter showed high occurrence around the mean. In the second scenario, the proposed filter showed the smallest deviation of position estimation error whilst the other filters showed larger deviation with larger maxima. In the case of iteration time, the proposed filter showed a much smaller deviation compared to PGM filters as shown in Fig. \ref{fig:exp_rosbag_results} (b) and (d). This indicates that the proposed filter shows consistency of good performance and robustness throughout the tested scenarios.

\section{Conclusion}\label{conclusion}
A new GMF was proposed with the addition of MI bounding and scaled sample covariance for resampling. The proposed filter was tested in a numerical benchmark problem and showed enhanced performance over PF and PGM filters. Moreover, the proposed filter was tested in the two autonomous robotic radio signal source search scenarios with partial observability and compared with PF and PGM filters. The results showed the efficiency and versatility of the proposed method over other filters in the partially observable scenario. Also, the experimental results showed that the proposed filter can be applied to the online radio signal source search scenario with robust estimation. However, the tested scenarios consider only one measurement from one robot whilst multiple measurements can be considered to improve the observability. Constructing the joint belief using a data fusion approach with multiple measurements from multiple robots to improve estimation performance remains as future work. Additionally, introducing hypothesis reduction and component merging can also reduce computation and is considered as future work.



\balance
\bibliographystyle{IEEEtran}
\bibliography{ref}

\end{document}